\documentclass{article}
\usepackage{ijcai23}

\usepackage{times}
\usepackage{soul}
\usepackage{url}
\usepackage[hidelinks]{hyperref}
\usepackage[utf8]{inputenc}
\usepackage[small]{caption}
\usepackage{graphicx}
\usepackage{amsmath}
\usepackage{amsthm}
\usepackage{booktabs}
\usepackage{algorithm}
\usepackage{algorithmic}
\usepackage[switch]{lineno}

\usepackage{amssymb}
\usepackage{listings}
\usepackage{xcolor}
\usepackage[T1]{fontenc}
\usepackage[scaled]{beramono}
\usepackage{stmaryrd}
\usepackage{mdframed}
\usepackage{booktabs}

\hypersetup{
    colorlinks=false,
    }

\newtheorem{defn}{Definition}

\newcommand{\alist}{alist}

\newcommand{\llb}{\llbracket}
\newcommand{\rrb}{\rrbracket}
\newcommand{\sem}[1]{\llb #1 \rrb}
\newcommand{\lrangle}[1]{\langle #1 \rangle}
\newcommand{\mc}[1]{\mathcal{#1}}
\newcommand{\question}{}

\colorlet{punct}{red!60!black}
\definecolor{background}{HTML}{f3fbff} 
\definecolor{rulecolor}{HTML}{d7ebf8} 
\definecolor{delim}{RGB}{20,105,176}
\colorlet{numb}{magenta!60!black}
\definecolor{keycolor}{HTML}{569CD6}

\lstdefinestyle{jsonstyle}{
    basicstyle=\normalfont\ttfamily\small, 
    numbersep=8pt,
    showstringspaces=false,
    breaklines=true,
    frame=lines,
    rulecolor=\color{rulecolor},
    backgroundcolor=\color{background},
    keywords={h,v,s,p,o,t},
    keywordstyle=\color{keycolor},
    literate=
     *{0}{{{\color{numb}0}}}{1}
      {1}{{{\color{numb}1}}}{1}
      {2}{{{\color{numb}2}}}{1}
      {3}{{{\color{numb}3}}}{1}
      {4}{{{\color{numb}4}}}{1}
      {5}{{{\color{numb}5}}}{1}
      {6}{{{\color{numb}6}}}{1}
      {7}{{{\color{numb}7}}}{1}
      {8}{{{\color{numb}8}}}{1}
      {9}{{{\color{numb}9}}}{1}
      {:}{{{\color{punct}{:}}}}{1}
      {,}{{{\color{punct}{,}}}}{1}
      {\$}{{{\color{punct}{\$}}}}{1}
      {?}{{{\color{punct}{?}}}}{1}
      {\{}{{{\color{delim}{\{}}}}{1}
      {\}}{{{\color{delim}{\}}}}}{1}
      {[}{{{\color{delim}{[}}}}{1}
      {]}{{{\color{delim}{]}}}}{1},
}

\title{ALIST: Associative Logic for Inference, Storage and Transfer.
\\
A Lingua Franca for Inference on the Web
}

\author{
Kwabena Nuamah
\and
Alan Bundy
\affiliations
University of Edinburgh, United Kingdom\\
\emails
\{k.nuamah, a.bundy\}@ed.ac.uk
}

\begin{document}

\maketitle

\begin{abstract}
    Recent developments in support for constructing knowledge graphs have led to a rapid rise in their creation both on the Web and within organisations. Added to existing sources of data, including relational databases, APIs, etc., there is a strong demand for techniques to query these diverse sources of knowledge. While formal query languages, such as SPARQL, exist for querying some knowledge graphs, users are required to know which knowledge graphs they need to query and the unique resource identifiers of the resources they need. Although alternative techniques in neural information retrieval embed the content of knowledge graphs in vector spaces, they fail to provide the representation and query expressivity needed (e.g. inability to handle non-trivial aggregation functions such as regression). We believe that a lingua franca, i.e. a formalism, that enables such representational flexibility will increase the ability of intelligent automated agents to combine diverse data sources by inference. 
    Our work proposes a flexible representation (\textit{alists}) to support intelligent federated querying of diverse knowledge sources. Our contribution includes (1) a formalism that abstracts the representation of queries from the specific query language of a knowledge graph; (2) a representation to dynamically curate data and functions (operations) to perform non-trivial inference over diverse knowledge sources; (3) a demonstration of the expressiveness of \textit{alists} to represent the diversity of representational formalisms, including SPARQL queries, and more generally first-order logic expressions.

\end{abstract}

\section{Introduction}
\label{introduction}
Knowledge graphs are being widely adopted for storing diverse forms of knowledge in different domains for different purposes. They range from large corporate knowledge graphs, used to support search engines (e.g. Wikidata\cite{vrandevcic2014wikidata}, Google Knowledge Graph \cite{singhal2012googleknowgprah} and DBPedia \cite{Bizer2009a}), to small personal knowledge graphs. Despite their rapid adoption, there is still a significant amount of fragmentation due to syntactic and semantic heterogeneity resulting from diverse knowledge representation formalisms, query languages and data storage mechanisms. This is to be expected given the different needs of users of these knowledge graphs. While some are interested in the ability to traverse graphs to determine the existence of connections between entities (e.g. link prediction tasks), others have strong requirements of the underlying ontology that provides semantics to the graph (e.g. types and cardinality). Given that these knowledge representation formalisms each have their own query language, it is often impossible to query across different formalisms without resorting to some custom program for the desired purpose. This creates a huge challenge in automatic data integration, a key component of the data engineering task in the data science automation pipeline \cite{de2021automating}.
It is particularly obvious in question answering over diverse web sources where using one query language such as SPARQL \cite{prud2008sparql} is insufficient to retrieve information from other sources. 

We provide a formalism that is expressive enough to represent queries of different types and serves as a lingua franca for inference, data integration and exchange purposes. This enables dynamic curation of retrieved knowledge into a common normal form without requiring knowlede bases (KBs) or data providers to adopt a new format. It also provides a structure for inferring new knowledge from sources in diverse formalisms and is compatible with existing web data exchange formats. 
In this paper, we discuss the \emph{alist\footnote{Previous use of the ALIST acronym has been limited to the \textit{association list} data structure. In this work, we impose additional properties on this data structure to induce a logic.} (Associative Logic for Inference, Storage and Transfer)} which help us to solve the above issues. 
We also demonstrate how the alist is useful in open-domain QA systems such as FRANK (Functional Reasoner for Acquiring New Knowledge) \cite{NuamahPhD2018,nuamah2016functional}. 


\section{Background}
\label{background}
Data \textit{triples} have long been used to represent knowledge. Especially in knowledge representations for  Linked Data \cite{bizer2009linked} and the Semantic Web \cite{berners2001semantic}, \textit{$\langle$subject, predicate, object$\rangle$} triples form the core of the Resource Description Framework (RDF) \cite{beckett2004rdf}. Several of these facts, usually with unique resource identifiers (URIs) for the items in the triple, are stored in triple stores or graphs to create ontologies or knowledge graphs. The SPARQL query language uses graph patterns \cite{sparqlGraphPatterns} to match triple patterns in a query to a subgraph of the RDF data in order to instantiate variables in the SPARQL query. Federated queries are answered by using data that is distributed across the web. SPARQL supports federated queries using the \textit{SERVICE} extension. However, query federation in SPARQL \cite{sparqlfederated2013} is limited to homogeneous (i.e. RDF) data sources. This makes it impossible to use the same representation and query language for data from relational databases, graph databases and a whole range of NoSQL data sources.

Generally, the task of data integration over heterogeneous data sources and formats has been a challenging one that lacks a clear formal process or representation for integration, query, and inference without ad hoc and manual methods. Recent reviews of data integration methods \cite{hendler2014data} highlight the human-intensive nature of this task and the lack of a simple and flexible format for curating data, not to mention the discovery and inference difficulties. Efforts to use machine learning (ML) to tackle this problem, such as \emph{Tamr} \cite{stonebraker2018data}, also acknowledge practical difficulties in training such models. Crucially, they fail to tackle the knowledge representation issue necessary for integration, focusing instead on specific workflows for ML tasks.


\section{New Formalism}
\label{new_formalism}
To tackle the aforementioned challenges, we create and formalise a lingua franca representation with the desired properties outlined below.

\subsection{Desiderata}
\label{desiderata}
In our new representation, we seek to satisfy the following requirements. First, we need a representation that is flexible enough for knowledge (data) and query representation, inference, and the exchange of knowledge between systems. We also need this representation to be web-friendly, allowing for easy use in JSON-based (Javascript Object Notation)\footnote{\url{https://www.json.org}} APIs for web systems. Finally, we require a formalism that is future-proof for neurosymbolic inference, allowing for both a symbolic treatment of the representation as well as a neural network-based embeddings of the representation and inferences that use it \cite{nuamah2021deepalgo}. While this specific task is outside the scope of this work, it influences our choice of data format such that it can readily be adopted for that task.

It is worth noting that our objective is not to replace existing data formalisms or query languages, but to provide one that seamlessly supports our need for dynamic data integration and automated inference over diverse KBs while being immediately usable on the web.

\subsection{Alist Semantics}
\label{alist_semantics}
We propose the \textit{alist} (\textbf{A}ssociative \textbf{L}ogic for \textbf{I}nference, \textbf{S}torage and \textbf{T}ransfer) as a formalism that meets the above criteria. Below, we define concepts and terminology that we use.


 \begin{defn}[Alist]
 \label{def-alist}
 An alist is defined recursively as a set of attribute-value pairs $\langle x: y \rangle$ to represent a question or fact such that $x \in  \{\mathcal{F} | \mathcal{O} | \mathcal{M} | variable\}$, where $\mathcal{F}$ is a set of functional attributes, $\mathcal{O}$ is a set of object-level attributes and $\mathcal{M}$ is a set of meta-level attributes (see definition 2,3 and 4); $y \in \{\textbf{C} | \textbf{v} | \mathcal{A}\}$ where $\textbf{C}$ are constant values and $\textbf{v}$ are variables (\S\ref{variables}), and $\mathcal{A}$ is an alist.
\end{defn}

It is worth noting that this recursive definition of an alist enables the construction of nested alists for more complex questions. Alists which do not contain nested alists are called \textbf{\textit{simple alists}}.

In the definitions that follow, we write the attribute names and their respective shortened symbols as $attribute(symbol)$ and express examples as:
\\
\\
\textit{Example 1: Simple alist (Tokyo was the capital of Japan in 1960):}
\begin{lstlisting}
{s:Japan, p:capital, o:Tokyo, t:1960}
\end{lstlisting}
Queries are expressed using variables (prefixed with $?$ or $\$$).
\\
\textit{Example 2: Simple alist query (Capital of Japan in 1960?):}
\begin{lstlisting}
{h:value, v:?x, s:Japan, p:capital, o:?x, t:1960}
\end{lstlisting}
\textit{Example 3: Nested query (Country with the highest gdp?):}
\begin{lstlisting}
{h:max, v:$y, s:?x, p:gdp, o:$y, 
   $y: {s:?z, p:type, o:country}}
\end{lstlisting}
More examples of alists are provided in \ref{tab:alist-lcquad} and in the supplementary data.

\begin{defn}[Functional Attributes, $\mathcal{F}$]
\label{def-functional-attr}
 These are types of attributes that define the functional operation on an alist. 
 $$\mathcal{F} = \{operation(h), operation\_variable (v)\}$$
\end{defn}
Operations $h$ are denoted by labels that reference functions from libraries of arithmetic, statistical or string functions. Custom operations can also be used, allowing, for instance, pre-trained neural networks to be used as operations.
The default operation of an alist is \textit{value}, which is an identity function on the operation variable. In example 3 above, $h=max$ and $v=\$y$, which gives a functional interpretation of the alist as $max(\$y)$.

\begin{defn}[Object-Level Attributes, $\mathcal{O}$]
\label{def-object-attr}
 These attributes represent the object information that captures the meaning of a fact or question. 
 \begin{align*}
     \mathcal{O} = & \{subject(s), property(p), object(o), time(t), \\
                   & ~~~location(l) \}
 \end{align*}
\end{defn}
These are comparable to the data triples found in RDF and other relational representations. 
The optional time and location attributes allow one to concisely express when and where (respectively) the fact expressed in the $\langle$\emph{s,p,o}$\rangle$ holds true.

\begin{defn}[Meta-Level Attributes, $\mathcal{M}$]
\label{def-metal-attr}
 These attributes capture meta-information about the question or fact as well as meta-data that is generated when variables are instantiated via retrieval or inference.
 \begin{align*}
     \mathcal{M} = & \{explanation(x), context(c), uncertainty(u), \\
                   & ~~~data\_source(d), \dots \}
 \end{align*}
 \end{defn}
 They include explanations of inferred values, contexts such as user preferences, uncertainty in variable instantiations, data sources used, etc.
 Meta-level attributes are also the mechanism through which alists can be extended to capture new kinds of meta-information during inference.


\begin{defn}[Alist Scope]
\label{def-alist-scope}
 The scope of an alist $\mathcal{A}$ is defined by the set of attributes within it. The \textbf{local scope} consists of all attributes in $\mathcal{A}$ except attributes in its nested child or parent alist. The \textbf{global scope} includes all attributes of $\mathcal{A}$, including those from its nested (or parent) alists.
\end{defn}

\subsection{Variables}
\label{variables}
Variables in alists are attributes that can be bound or instantiated to a specific value. Variables are defined in the \emph{local scope} of an alist. Three types of variables are used in an alist. Figure \ref{fig:alist_variables} illustrates the different kinds of variables defined below. 
\begin{defn}[Projection Variable]
 A variable whose bound value is returned (or extracted) from an alist. Such variables are prefixed with $?$, e.g., $?x$. Each alist can have only one projection variable in its scope.
\end{defn}

\begin{defn}[Auxiliary Variable]
 A variable that is only used within an alist and whose value is not projected from the alist. The variable name is prefixed with $\$$, e.g., $\$z$.
\end{defn}

\begin{defn}[Operation Variable]
 A variable or list of variables assigned to the operation variable attribute $v$ which serves as the argument of the alists's operation $h$. 
 Formally, we define an operation variable in an attribute-value pair of an alist as as $\langle v : x_1 ~ | ~[x_1, ..., x_n]  \rangle$, where $x_i$ is a variable in the local scope of the alist. Further,
 \begin{itemize}
     \item both projection and auxiliary variables can be used as operation variables;
     \item all alists must have an operation variable, a projection variable or both;
     \item if no projection is specified in an alist, the operation variable is projected from the list;
     \item if no operation variable is provided in an alist, the projection variable is set as the operation variable.
 \end{itemize}
 \end{defn}
\noindent The following alists are therefore equivalent:
\begin{lstlisting}
{h:value, v:?x, s:Japan, p:capital, o:?x, t:1960}
{h:value, s:Japan, p:capital, o:?x, t:1960}
{s:Japan, p:capital, o:?x, t:1960}
\end{lstlisting}

\begin{figure}[t]
    \begin{center}
    \includegraphics[width=0.45 \linewidth]
    {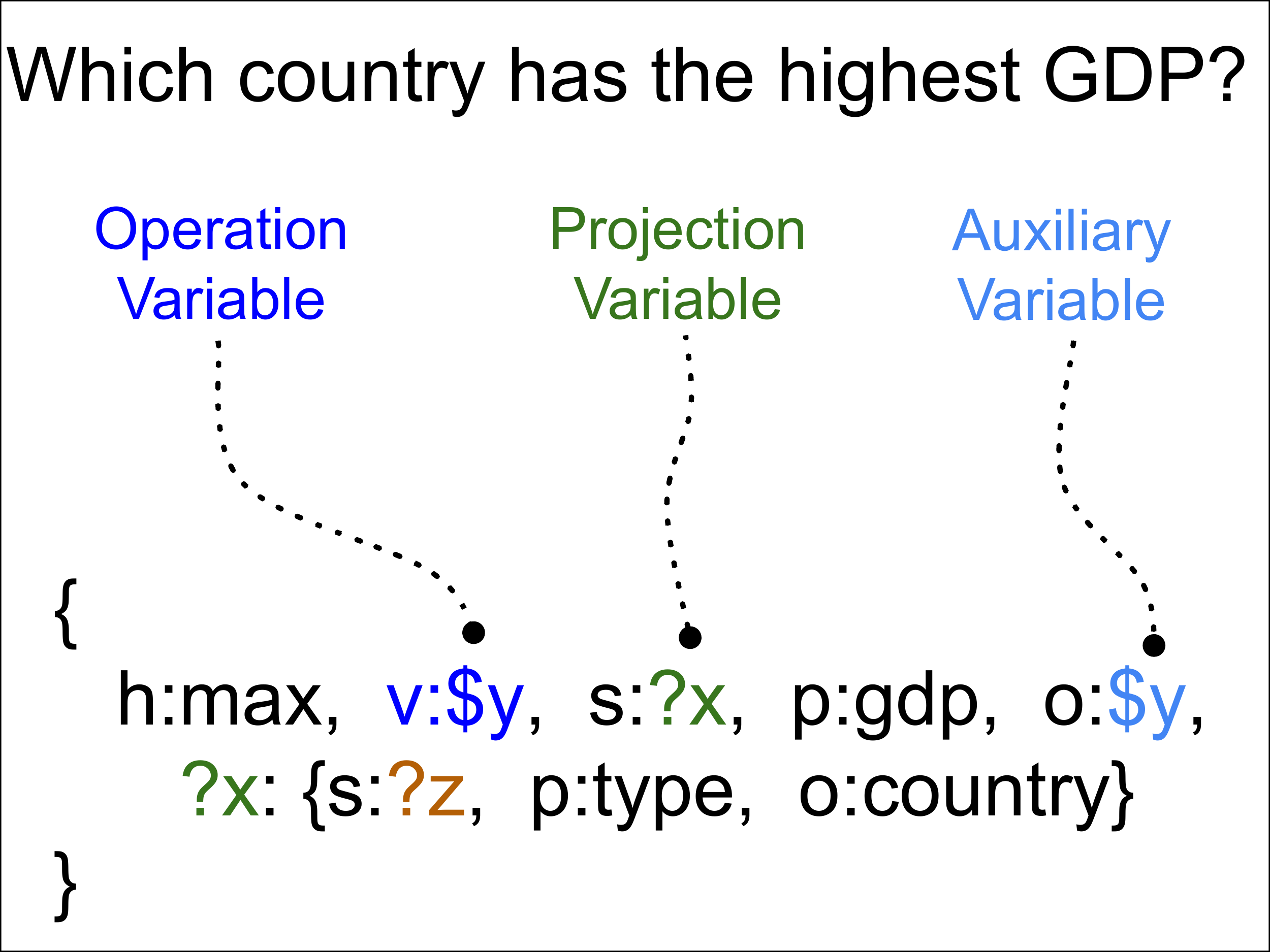}
    \end{center}
    \caption{Showing projection, auxiliary and operation variables in an alist.}
    \label{fig:alist_variables}
    \end{figure}

\textbf{Resolving} an alist is the process of instantiating its variables. \emph{Fully resolved} alists have all their variables instantiated, while \emph{partially resolved} alists have at least one uninstantiated variable.

\section{Formalizing Alists}
\label{formalizing_alists}
In the sections below, we formalize the alist with the goal of demonstrating its semantics and expressivity.

\subsection{Semantics}

We describe the semantics of the alist over a set of knowledge bases (KBs) and inference environment. As previously mentioned, we distinguish between \textit{object-level} and \textit{meta-level} components. 
Object-level components are data items obtained from a set of KBs $\mathcal{K}$, a set of KBs.
Meta-level components, including user context and uncertainty, and attribute and function names are considered part of an inference system's \textit{environment}, $\eta$.

We denote the semantics of an alist $\mathcal{A}$ by $\sem{\mathcal{A}}$. 
This is a function that takes a set of KBs, $\mathcal{K}$, 
and an environment, $\eta$, as inputs and produces the output $\sem{\mathcal{A}}_\mc{K,\eta}$ which are facts obtained by resolving $\mathcal{A}$ using data in $\mathcal{K}$ under constraints specified in $\eta$.
We define the semantics of components of an alist using the semantic function $\sem{\cdot}$ in figure \ref{semantics}.

\begin{figure}[h]
\begin{mdframed}
Let $\mathcal{A}$ be an alist, $a_i$ be attributes, $\phi$ be data values, and $\sem{\cdot}$ denote the semantics function.
\begin{align*}
&\sem{\phi}_\mc{K,\eta} =
\begin{cases}
	\phi \in  \mathcal{K}, & \text{for object-level values} \\
	\phi \in \eta, & \text{for meta-level values}
\end{cases} \\
&\sem{a_i}_\mathcal{\eta} = a_i \in \eta
\\ 
&\sem{\lrangle{a_i, \phi}}_\mathcal{K,\eta} = \lrangle{ \sem{\mc{A}_i}_\mc{\eta},  \sem{\phi}_\mc{K,\eta}} 
\\
&\sem{\mathcal{A}}_\mc{K,\eta} \equiv 
        \sem{ \big\{ \lrangle{a_1, \phi_1}, \dots,   
     	\lrangle{a_n, \phi_n} \big\}}_\mc{K,\eta} \\
&\sem{\big\{ \lrangle{a_1, \nu_1}, \dots,   
	              \lrangle{a_n, \phi_n} \big\}}_\mc{K,\eta} = \\
& ~~~~~~~~ \mathcal{A}\big\{ \sem{\lrangle{a_1, \phi_1}}_\mathcal{K,\eta}, \dots,   
	          \sem{\lrangle{a_n, \phi_n}}_\mathcal{K,\eta} \big\} \\
&\sem{\langle \mathcal{A}_1, \dots, \mathcal{A}_n \rangle}_\mc{K,\eta} = \langle \sem{\mathcal{A}_1}_\mc{K,\eta},
              \dots, \sem{\mathcal{A}_n}_\mc{K,\eta} \rangle, \\
& ~~\text{where} ~~ n > 0. 
\end{align*}

Given operation $\lambda$ and operation variable, $\nu$,  a functional interpretation of an alist over its decomposition (i.e. child alists):
\begin{align*}
&\sem{\mathcal{A} 
	 \big\{ 
	     \lrangle{h, \lambda}, 
	     \lrangle{v, \nu} 
	 \big\}, \dots }_\mc{W,\eta} = \\
& ~~~~~~~ ~~~~~~~ ~~~~~~ ~~~~~~~~~    
\lambda\big( \sem{\mathcal{A}_1[\nu]}_\mc{W,\eta}, \dots, 
 \sem{\mathcal{A}_n[\nu]}_\mc{W,\eta} \big) \\
&     ~~\text{where \textit{children}}(\mathcal{A}) = \lrangle{\mathcal{A}_1, \dots, \mathcal{A}_n}, \\
      & ~~h \in \eta 
        ~~\text{and} ~~ n > 0 
\end{align*}
Where $\mathcal{A}_i$s are formed from a decomposition, $\Delta$, of $\mathcal{A}$  such that:
\begin{align*}
\sem{\Delta(\mathcal{A}))}_\mc{W,\eta}  = \sem{\langle \mathcal{A}_1, \dots, \mathcal{A}_n \rangle}_\mc{W,\eta}
\end{align*}

\end{mdframed}
\caption{Alist Semantics} \label{semantics}
~
\end{figure}

\subsection{First-Order Logic to Alists}
\label{fol}
In this section, we discuss the expressivity of alists and show how they can represent first-order logic. 
The translation preserves both the logical and functional properties of alists and FOL constructs.
We begin with a case-based translation of FOL expressions to alists.

\begin{defn}[FOL BNF]
To define the case of translating from FOL to alists, we will use the following FOL BNF\footnote{We omit $\implies$ as it can defined using $\vee$ and $\neg$.}

\begin{align*}
    &Formulae: & F := &  P(T_1, \dots, T_n) ~|~ F \wedge F ~|~ F \vee F ~|~ \neg F ~|~ \\
    &~         & ~    & \forall V ~F ~|~ \exists V ~F \\
    &Terms: ~~ & T := & C ~|~ V ~|~ G(T_1, \dots, T_n)
\end{align*}
where $P$, $G$, $C$ and $V$ are sets of predicate symbols, function symbols, constants and variables, respectively, and $n$ is a positive integer.
\end{defn}

\subsubsection{Propositions}
\begin{defn}[Propositional Alists] \label{def_propo_alist}
Suppose the proposition is $P(T_1, \dots, T_n)$. An alist is created with  $n+1$ attribute/value pairs. Using the \textit{property} attribute $P$ and subsequent attributes $Arg_i$ for $1\leq i \leq n$, we get

\begin{lstlisting}[mathescape=true]
{p:P, $Arg_1$:$T_1$, $\dots$, $Arg_n$:$T_n$}
\end{lstlisting}
where the $Arg_i$ could be given more meaningful names if these are available. 
\end{defn}

This definition only deals with the Datalog sub-logic of FOL, in which the function case $G(T_1, \dots, T_n)$ from the BNF of terms is omitted. To deal with the general case, such non-nullary functions can themselves be translated to alists as in Definition \ref{def_func_alist}.

\begin{defn} [Functional Alists]\label{def_func_alist}
Let the function be $G(T'_1, \dots, T'_n)$. By analogy with the translation of propositions to alists in Definition 2, each nested non-nullary function of the form is translated to an alist as follows.

\begin{lstlisting}[mathescape=true]
{p:G, $Arg_1$:$T'_1$, $\dots$, $Arg_n$:$T'_n$}
\end{lstlisting}
\end{defn}
where $T_i$ is renamed to $T'_i$ to avoid clashes with the arguments of $P$ in Definition \ref{def_propo_alist}. Note that this translation process must proceed as deeply as functions are nested within propositions.

As an example\footnote{
Note that we do not assume that the property (or predicate) "\textit{starts with}" exists in a KB. We only use the surface text in the alist and leave the alignment of that term with the appropriate property label from KBs as a separate retrieval task. The same applies to entity labels used in other object-level attribute values.
}, 
consider the following query: \textit{``Name a military rank that starts with the letter M''}. This might be represented as the nested alist:
\begin{lstlisting}
{p:startswith, s:?x, o:M, 
 ?x: {p:type, s:?y, o:"military rank"}}
\end{lstlisting}
normalised into two alists:
\begin{lstlisting}[mathescape=true]
{p:startswith, s:?x, o:M, ?x: $C$} 
\end{lstlisting}
and
\begin{lstlisting}[mathescape=true]
{p:type, s:?y, o:"military rank"}
\end{lstlisting}
where $C$ is the value that instantiates the projection variable $?y$ in the second alist.

\subsubsection{Connectives}
\label{connectives}
FOL encompasses the construction of compound formulae in which propositions are combined with connectives. This, in effect, creates alists with compound goals. 

\begin{defn}[Conjunction and Disjunction] \label{def_conj_disj}
Conjunctions and disjunctions are constructed by using the AND and OR operations (respectively) as values for the \textit{h} attributes of the alist and nesting the conjuncts (or disjuncts) in the alist.
\end{defn}

Consider the question, \textit{``What is the cause of death and the place of death of John Denver?''}. A conjunctive alist can be constructed as:
\begin{lstlisting}
{h:AND, v:[$x,$y], 
 $x: {p:"cause of death", s:"John Denver", o:$x},
 $y: {p:"place of death", s:"John Denver", o:$y}}
\end{lstlisting}

Similarly, a disjunctive alist can be constructed for the question \textit{``What is either the cause of death or the place of death of John Denver?''} by replace \textit{AND} with \textit{OR} in the above alist.

\begin{defn}[Negation] \label{def_negation}
Negation of an alist ($\neg \mathcal{A}$) amounts to showing that all possible instantiation to variables in $\mathcal{A}$ makes the alist false.
\end{defn}
Three approaches are possible:
\begin{itemize}
    \item Treat negation as a first-class object using the \emph{NOT} operation. For instance to get entities that are not the capital of UK: 
    \begin{lstlisting}
{h:NOT, v:?x, p:"capital", s:"UK", o:?x}
    \end{lstlisting}
    This works under the closed world assumption (CWA), which is realistic for a database with narrow scope, but is not realistic when considering the whole Internet, which is incomplete and is not amenable to an exhaustive search.

    \item Use negation as failure, i.e., to show $\neg \mathcal{A}$ that was false, fail to prove A in an exhaustive search. Again, this only works under the CWA. Note that this approach can include free variables, if we fail to find any value that makes it true.

    \item Prove that a function has a different result, i.e., to show
    $f(x) = c$, where $c$ is a constant, prove instead that $\neg f(x) = c'$, and that $c \neq c'$. 
    This might even be extended to non-constant answers, e.g., prove
$\forall x. c(x) \neq c'(x)$. Note that by $f(x) = c$, we mean some alist $\mathcal{A}(x, c)$, where we have additional information
that, given $x$, then $c$ is unique, i.e., that it is a function onto $c$.
\end{itemize}

For example, suppose the query was: \textit{``Was Santa Cruz not John Denver’s place of death?''}
We might instead try to find his actual place of death and show that it wasn't \textit{Santa Cruz}. So, the goal would be: 
\begin{lstlisting}
{p:"cause of death", s:"John Denver", o:?x}
\end{lstlisting}
Suppose $?x$ is instantiated to \textit{Monterey Bay}, and that we can also prove \textit{Santa Cruz} $\neq$ \textit{Monterey Bay},
then we can assume that the goal is false so that the original query can be confirmed. Note that we've swapped one
negative goal for another - but a simpler one. We could show that \textit{Santa Cruz} $\neq$ \textit{Monterey Bay}, for instance, by
adopting the unique name assumption or by finding their coordinates and showing that they were unequal.

\subsubsection{Quantifiers} 
\label{quantifiers}
Resolution deals with quantifiers via Skolemisation, in which universal variables become free variables and existential variables become Skolem functions. Note, however, that goals in resolution are first negated before
being Skolemised. Since we do not negate queries, we need to employ a dual form of Skolemisation, in which it is universal variables that become Skolem functions and existential ones that become free variables
\cite{bundy_1983}[\S15.3].

From a pragmatic viewpoint, the justification of dual Skolemisation is that we want to instantiate the existential
variables to the values that answer the query, which representing them as free variables permits. Universal variables,
on the other hand, should not be instantiated, which risks answering only a special case of the query. Representing
them as ground terms, e.g., Skolem constants, prevents them from being instantiated. 

\begin{defn}[Dual Skolemisation] \label{def_dual_skolem}
Assume, without loss of generality, that a formula is in prenex form, i.e.,
that all quantifiers are at the front whose scope consists only of alists and connectives. We can formalise such a prenex formula\footnote{It is a normal part of the conversion to clausal form, to convert any FOL formula into an equivalent prenex formula, so we assume this has been done.} as:
$$Q_1x_1, \dots, Q_nx_n.\phi(x_1, \dots, x_n)$$
where each $Q_i$ is either $\forall$ or $\exists$ for $1\leq i \leq n$, and $\phi(x_1, \dots, x_n)$ is quantifier free, so consists only of connectives applied to alists.

Let $\mathcal{V}$ be a list of free variables whose initial value is the empty list $[~]$. We define dual skolem recursively as follows: 
\begin{align*}
&dual\_skolem(\phi(t_1, \dots, t_n),\mathcal{V})  = \phi(t_1, \dots, t_n) \\
&dual\_skolem(\forall x_1, Q_1x_1, \dots, Q_nx_n.\phi(x_1,x_2, \dots, x_n),\mathcal{V}) \\
& = dual\_skolem(Q_2x_2, \dots, Q_nx_n.\phi(sk_1(\mathcal{V}), x_2, \dots, x_n), \mathcal{V}) \\
&dual\_skolem(\exists x_1, Q_1x_1, \dots, Q_nx_n.\phi(x_1,x_2, \dots, x_n),\mathcal{V}) \\
& = dual\_skolem(Q_2x_2, \dots, Q_nx_n.\phi(x_1, x_2, \dots, x_n),[x_1| \mathcal{V}])
\end{align*}
where $sk_1(\mathcal{V})$ is a new Skolem function applied to the free variables in $\mathcal{V}$.
\end{defn}

For example, in the question: \textit{``What will be the UK's annual population over the next decade?''}, could be translated in the alist query:
\begin{lstlisting}[mathescape=true]
$\forall x.\exists ?y$ {s:UK, p:population, o:$?y$, t:$x$,
         $x$: {s: $x$, p:range, o:[2022,...,2031]}}
\end{lstlisting}
which would be dual skolemised to: 
\begin{lstlisting}[mathescape=true]
{s:UK, p:population, o:$?y$, t:$sk$,
  $sk$: {s: $sk$, p:range, o:[2022,...,2031]}}
\end{lstlisting}
where $sk$ is a new Skolem constant. Since, in this example, $sk$ has a finite and small range, we can call a separate query for each of the possible values of $sk$, e.g.,
\begin{lstlisting}[mathescape=true]
{s:UK, p:population, o:$?y$, t:2025}
\end{lstlisting}

\section{Using Alists}
We now show how alists achieve our goals for inference, dynamic curation (i.e. storage) and data exchange (i.e. transfer).
\label{using_alists}

\subsection{Dual Data and Functional Views of Alists}
\label{dual_role}
Object-level attributes in an alist define the domain data that is captured in the alist (thus, data about properties of entities or real-world concepts). 
However,  functional attributes (Def \ref{def-functional-attr}) allow us to express statements about the values of the object-level attributes and specify operations to perform on attributes in the alist.
Since values of attributes could also be functions (see Def \ref{def-alist}), this means that arithmetic or statistical functions (e.g. regression curves) can be assigned to an object-level attribute in an alist. 

\begin{figure}[t]
    \begin{center}
    \includegraphics[width=1.0 \linewidth]
    {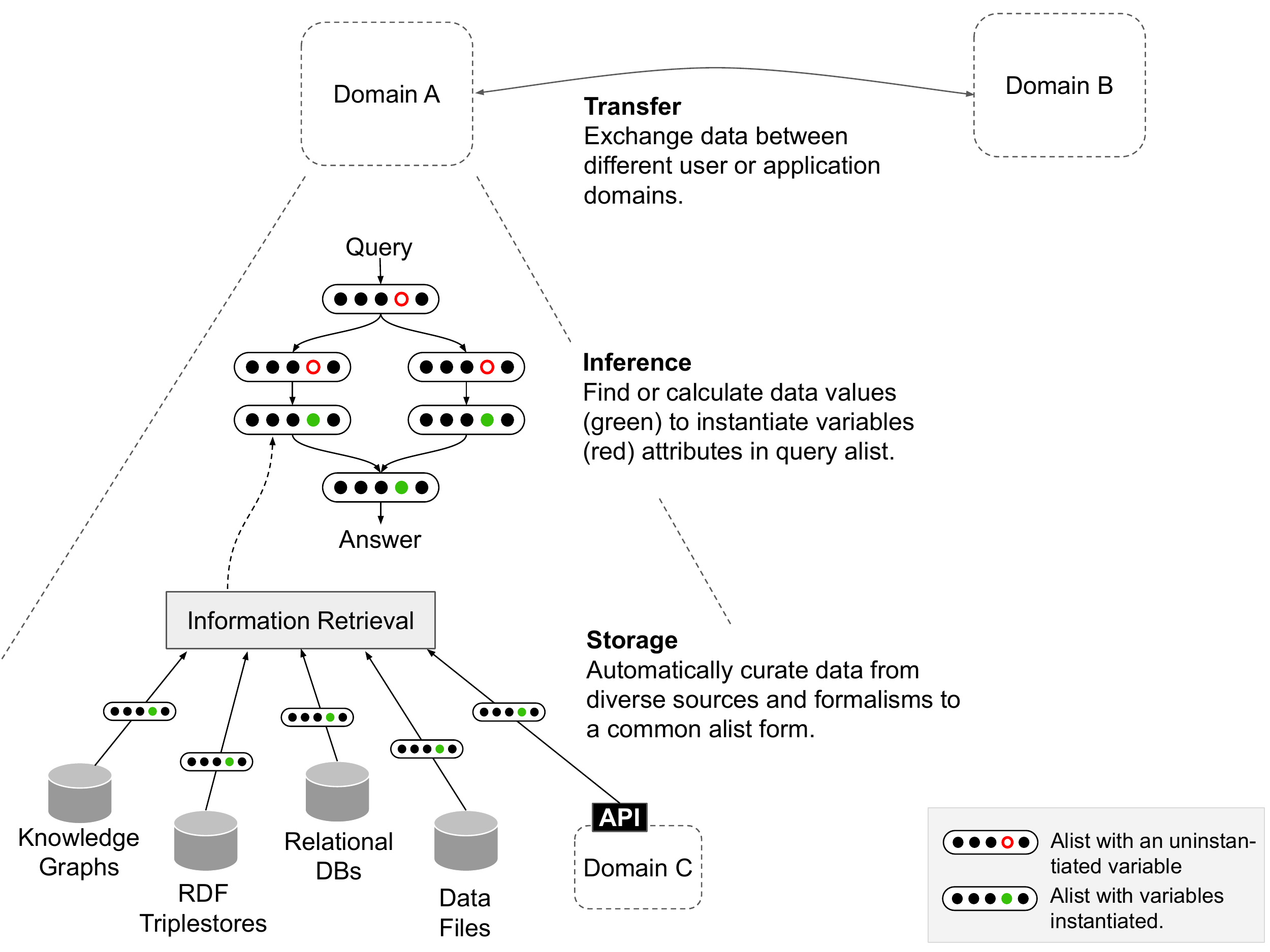}
    \end{center}
    \caption{Illustrating our three main uses of alists: inference, storage and transfer.}
    \label{fig:alist_illust}
    \end{figure}

Similarly, given that an alist can be interpreted as a function with the operation attribute as the function name and the operation variable as its operand, we can assign values of projection variables from one alist to values of attributes in another \alist.
This feature gives an alist its functional and nested query capabilities, and is also a significant difference between alists and RDF.
Next, we discuss how alists are used for inference as well as the flexibility we gain from using them. These are illustrated in figure \ref{fig:alist_illust}.  

\subsection{Alists for Inference}
\label{sec-inference}
The goal of inference is to resolve all relevant variables in an alist. To answer a query, relevant variables are instantiated by retrieving values from KBs or inferring values from existing data through the application of \emph{decomposition} and \emph{aggregation} operations.

\subsubsection{Decomposition}
\label{decomposition}
A map operation that takes as input an alist $\mathcal{A}$ (i.e. the parent alist) and a decomposition rule, $\tau$ and returns a list of new alists (called its child alists).
There are three main types of decomposition rules:
\begin{enumerate}
    \item \emph{Normalization:} an operation on an alist to decompose it into one or more simple alists. This is usually the initial decomposition step when resolving a nested alist.
    
    \item \emph{Partition:} an operation on an alist to partition one of its object-level attributes predominantly using the \textit{partOf} (or its inverse) relation. For example, a geospatial partition decomposes a place into its constituent and non-overlapping regions (e.g. a continent into its countries), while an object partition splits an object into its constituent parts.
    
    \item \emph{Sequence:} an operation on an alist to split the value of an object-level attribute into a sequence of values to create new child alists. In particular, there is an inherent order to the list of child alist. For instance, a temporal decomposition on the \textit{time attribute} is used to split the alist over a specified period of time (e.g. years in a decade). This is useful for performing regression over time to predict the value of the alist at specified time.
\end{enumerate}

\subsubsection{Aggregation}
\label{aggeregation} 
A reduce operation on a list of child alists to combine values into the parent alist, where the values to be aggregated are determined by the operation variable. Aggregation functions include arithmetic, statistical, string, etc., operations.

\subsubsection{Inference}
For a \emph{query} alist containing uninstantiated variables, inference is the task of instantiating the relevant variables needed to extract an answer from the alist using data from the available KBs. This is done by retrieving data from the KBs by translating simple alists to the native query language of those KBs.
However, since a query alist can be nested and not be immediately translated to a native query in its compound form, \textit{normalisation} (see \S\ref{decomposition}) is first used to decompose the alist into one or more simple alist.

Additionally, if a query on a normalised alist fails to retrieve any data from available sources, the alist can be decomposed to create new child alists that form new sub-goals to be solved. These child alists, once solved can be aggregated to solve the parent alist. This chain of deductive and inductive reasoning is used in the FRANK (Functional Reasoner for Acquiring New Knowledge) \cite{NuamahPhD2018,nuamah2016functional} question answering system. Such an inference mechanism is appropriate for reasoning over an open domain such as the internet to deal with the diverse, noisy and incomplete data on the web \cite{Bundy2018}. The inference mechanism also allows for the explicit handling of uncertainty in alists values and also provides an inference graph that can be used as a basis for generating explanations for users \cite{nuamah2020explainable}.

\subsection{Alist for Storage}
\label{sec-storage}
The expressiveness of alists, coupled with its simple representation makes it well-suited for automatically curating data from heterogeneous sources while retaining as much of the source semantics as possible.

The relational semantics of alists, based on its subject, predicate, and object attributes makes it straight forward to convert rdf triples to alists. More generally, it works for $\langle$head, relation, tail$\rangle$ triples in graph data.  Similarly, RDF reification can be used to translate other attributes of alists to rdf statements. Reification \cite{manola2004rdf} is the use of RDF statements to describe other RDF statements and to record information about when statements were made, who made them, or other provenance details. 
Consider the alist 
\begin{lstlisting}[mathescape=true]
{ s:$x$, p:$p$, o:$o$, $P_1:a_1$, $\dots$, $P_n:a_n$}$$
\end{lstlisting}
We can express this as a reified RDF statement using the following RDF triples: 
\begin{lstlisting}[mathescape=true]
  $\langle$ $q$, type, rdf:Statement $\rangle$
  $\langle$ $q$, rdf:subject, s $\rangle$
  $\langle$ $q$, rdf:predicate, p $\rangle$
  $\langle$ $q$, rdf:object, o $\rangle$
  $\langle$ $q$, $P_1$, $a_1$ $\rangle$
      $\dots$
  $\langle$ $q$, $P_n$, $a_n$ $\rangle$
\end{lstlisting}
where $q$ is the URI for the statement $\langle$s, p, o$\rangle$, and the $P_i$s are predicates representing the attribute URIs for the values $a_i$ for $1\leq i \leq n$, respectively.


\subsection{Alist for Transfer}
\label{sec-storage}
To effectively do web-scale inference, it is important to exchange data between diverse systems.
The attribute-value pair representation of alists makes it readily accessible to use in practical programming applications. In particular, alists can be represented as JSON, a popular data format for web applications and also an ECMA-International data interchange format (ECMA-404) \cite{ecma404_2017}. The attribute-value data structure (commonly implemented as associative arrays) is also available in most programming languages (e.g. Dictionaries in Python\footnote{\url{https://docs.python.org/3/tutorial/datastructures.html\#dictionaries}} and Maps in Java\footnote{\url{https://docs.oracle.com/javase/8/docs/api/java/util/Map.html}}).


\subsection{Implementation}
We provide an implementation of alists and an inference mechanism built using alists to dynamically answer queries that require data from Wikidata (in SPARQL/RDF), the World Bank (\url{https://data.worldbank.org/}) (in JSON via an API) and MusicBrainz (\url{http://musicbrainz.org}) (in JSON via an API). We provide alist queries of the following types: simple alist queries, nested queries, prediction queries,  No data is stored locally, forcing our system to query the three data sources using the same alist. Our code provides an interface to the data sources and translates simple alist queries to native queries of these sources. We also convert query results from these sources to alists.  
Code (in Python) is attached as supplementary material.


\section{Discussion}
\label{discssion}
Our initial implementation demonstrates that it is feasible to have a flexible formalism to model data and queries using a simple representation such that it facilitates the automatic integration of data in different formats while supporting inference to answer complex questions.
However, the alist is not meant to replace existing knowledge representation formalisms or query languages, e.g. SPARQL, SQL, Cypher \cite{francis2018cypher}, etc., nor is it meant to curate the whole web. Instead, it is aimed at complementing them to provide a domain- and platform-independent representation for query, inference, storage and transfer of facts between diverse knowledge providers. Also, unlike federated or distributed queries, our approach can use heterogeneous data sources with each one using a different query language or data representation. Dynamic curation also eliminates the need to convert the different knowledge bases to one format prior to using them. Not only is this impractical at web scale, but also the data is constantly changing, making such frequent wholesale conversions very expensive.  
Applications such as question-answering systems or fact-checking systems can benefit from such dynamic curation over diverse sources. Additionally, this is a step towards tackling the challenge of automated data integration for tasks in AutoML and AutoAI \cite{de2021automating}. The lightweight and flexible representation allow for new attributes to be added to the alist as needed for the specific data integration or inference task.


Although theoretically, any alist can be described in RDF using RDF reification, there are practical reasons why such reification is not best suited for our needs. Primarily, beyond RDF, reification is not uniformly available in other KB representations, particularly, those with a very lightweight schema. 

There are, however, limitations to our approach. For instance, the simple alist used to query data means that complex queries (e.g. those with joins) are not efficiently executed. We will explore this in future work.

\section{Conclusion}
As the volume of rich data from diverse sources increases, there is the need to accommodate these diverse representations for use in automated and intelligent systems in applications such as question answering. Our work is focused on providing a novel formalization of alists that offers such flexibility. We demonstrated its use for representing both queries and data, its use in inference (via decomposition and aggregation operations) and its appropriateness for exchanging information in distributed systems. We also demonstrated its expressiveness to represent SPARQL queries and, more generally, first-order logic expressions.

As future work, we will use this representation as a basis for embedding queries and data for use in neuro-symbolic systems. The dense alists can then be used to train models for decomposing and aggregating alists. This will be a step towards realising the goal of creating an inference system that supports symbolic reasoning while leveraging neural networks to learn and optimize its inference process.

\begin{table}[h]
\centering
\begin{tabular}{p{1.0\linewidth}}
    \toprule
       Alist examples from the LC-QuAD question types \\
    \midrule
        
{\textit{Single Fact:}}
\question{Who is the screenwriter of Mr. Bean?}
\begin{lstlisting}
{h:value, v:?x, s:"Mr. Bean", p:screenwriter, o:?x}
\end{lstlisting}
\\
{\textit{Multi-Fact:}}
\question{What is the name of the sister city tied to Kansas City, which is located in the county of Seville Province?}
\begin{lstlisting}
{h:value, v:?x, s:?x, p:"sister city", 
 o:"Kansas City", ?x: {s:?y, p:location, 
   o:"Seville Province"}}
\end{lstlisting} 
\\
{\textit{Facts with Qualifiers:}}
\question{What is the venue of Barak Obama's marriage?}
\begin{lstlisting}
{h:value, v:?y, s:$x, p:venue, o:?y, 
  $x: {s:"Barak Obama", p:marriage, o:?z}}
\end{lstlisting}
\\
{\textit{Boolean:}}
\question{Did Breaking Bad have 5 seasons?}
\begin{lstlisting}
{h:equal, v:[$x,$y], 
  $x: {h:count, v:?x, s:"Breaking Bad", p:seasons, o:?x},
  $y: 5}
\end{lstlisting}
\\
{\textit{Ranking:}}
\question{What is the binary star which has the highest colour index?}
\begin{lstlisting}
{h:rank, v:[$x,1], s:?y, p:"color index",o:$x,
  ?y: {h:value, v:?z, s:?z, p:type, o:"binary start"}}
\end{lstlisting}
\\
\bottomrule
\end{tabular}
\caption{Examples showing how different kinds of questions can be represented in an alist using 5 of the query types in the LC-QuAD dataset.}
\label{tab:alist-lcquad}
\end{table}

\appendix
\section*{Supplementary Data}
We provide supplementary data to evidence our claims:
\begin{enumerate}
    \item Alist representation for the 10 question types in the LC-QuAD dataset.
    \item Code in Python on an implementation of the alist together with an interface to Wikidata, the WorldBank dataset and MusicBrainz. It also contains an implementation of the FRANK inference method that uses decomposition an aggregation operations. We test this with 10 alist examples in the script provided.   
\end{enumerate}



\bibliographystyle{named}
\bibliography{alist}

\end{document}